\documentclass[sigconf]{acmart}

\AtBeginDocument{%
  }

\setcopyright{acmlicensed}
\copyrightyear{2025}
\acmYear{2025}
\acmDOI{XXXXXXX.XXXXXXX}

\acmConference[ACM MM ’25]{Make sure to enter the correct conference title from your rights confirmation email}{10/27-10/31}{Dublin, Ireland}
\acmISBN{978-1-4503-XXXX-X/18/06}

\acmSubmissionID{3305}
\usepackage{times}
\usepackage{epsfig}
\usepackage{graphicx}
\usepackage{amsmath}

\usepackage{booktabs}
\usepackage{commath}
\usepackage{mathtools}
\usepackage{enumerate}
\usepackage{enumitem}
\usepackage{xcolor}
\usepackage{soul}

\usepackage{caption}
\usepackage{subcaption}
\usepackage{multirow}
\usepackage[font=small,skip=3pt]{caption}

\usepackage{algorithm}       
\usepackage{bm}              
\usepackage{algpseudocode}
\usepackage{wasysym}

\usepackage[capitalize]{cleveref}
\crefname{section}{Sec.}{Secs.}
\Crefname{section}{Section}{Sections}
\Crefname{table}{Table}{Tables}
\crefname{table}{Tab.}{Tabs.}

\definecolor{amethyst}{rgb}{0.6, 0.4, 0.8}
\definecolor{mygray}{gray}{0.5}

\usepackage{xspace}
\makeatletter
\DeclareRobustCommand\onedot{\futurelet\@let@token\@onedot}
\def\@onedot{\ifx\@let@token.\else.\null\fi\xspace}

\makeatother

\renewcommand\footnotetextcopyrightpermission[1]{}

\begin{document}

\title{NavigScene: Bridging Local Perception and Global Navigation for Beyond-Visual-Range Autonomous Driving}



\renewcommand{\shortauthors}{Peng, et al.}

\author{Qucheng Peng}
\authornote{This work was done during Qucheng Peng's internship at XPENG Motors' Silicon Valley office in CA, USA.}
\email{qucheng.peng@ucf.edu}
\affiliation{%
  \institution{Center for Research in Computer Vision, University of Central Florida}
  \city{Orlando}
  \state{FL}
  \country{USA}
}

\author{Chen Bai}
\email{chenbai@xiaopeng.com}
\affiliation{%
  \institution{XPENG Motors}
  \city{Santa Clara}
  \state{CA}
  \country{USA}
 }

\author{Guoxiang Zhang}
\affiliation{%
  \institution{XPENG Motors}
  \city{Santa Clara}
  \state{CA}
  \country{USA}
 }

\author{Bo Xu}
\affiliation{%
  \institution{XPENG Motors}
  \city{Santa Clara}
  \state{CA}
  \country{USA}
 }

\author{Xiaotong Liu}
\affiliation{%
  \institution{XPENG Motors}
  \city{Santa Clara}
  \state{CA}
  \country{USA}
 }

\author{Xiaoyin Zheng}
\affiliation{%
  \institution{XPENG Motors}
  \city{Santa Clara}
  \state{CA}
  \country{USA}
 }

\author{Chen Chen}
\email{chen.chen@crcv.ucf.edu}
\affiliation{%
  \institution{Center for Research in Computer Vision, University of Central Florida}
  \city{Orlando}
  \state{FL}
  \country{USA}
}

\author{Cheng Lu}
\email{luc@xiaopeng.com}
\affiliation{%
  \institution{XPENG Motors}
  \city{Santa Clara}
  \state{CA}
  \country{USA}
 }

\renewcommand{\shortauthors}{Peng et al.}

\begin{abstract}
Autonomous driving systems have made significant advances in Q\&A, perception, prediction, and planning based on local visual information, yet they struggle to incorporate broader navigational context that human drivers routinely utilize. We address this critical gap between local sensor data and global navigation information by proposing NavigScene, an auxiliary navigation-guided natural language dataset that simulates a \emph{human-like} driving environment within autonomous driving systems. Moreover, we develop three complementary paradigms to leverage NavigScene: (1) Navigation-guided Reasoning, which enhances vision-language models by incorporating navigation context into the prompting approach; (2) Navigation-guided Preference Optimization, a reinforcement learning method that extends Direct Preference Optimization to improve vision-language model responses by establishing preferences for navigation-relevant summarized information; and (3) Navigation-guided Vision-Language-Action model, which integrates navigation guidance and vision-language models with conventional driving models through feature fusion. Extensive experiments demonstrate that our approaches significantly improve performance across perception, prediction, planning, and question-answering tasks by enabling reasoning capabilities beyond visual range and improving generalization to diverse driving scenarios. This work represents a significant step toward more comprehensive autonomous driving systems capable of navigating complex, unfamiliar environments with greater reliability and safety.
\end{abstract}

\ccsdesc[500]{Computing methodologies~Computer vision tasks}
\ccsdesc[500]{Information systems~Multimedia databases}

\keywords{Vision-language Model, Multi-modal Learning, Autonomous Driving}



\maketitle

\section{Introduction}

\begin{figure}[!ht]
    \centering
    \includegraphics[width=1.0\linewidth]{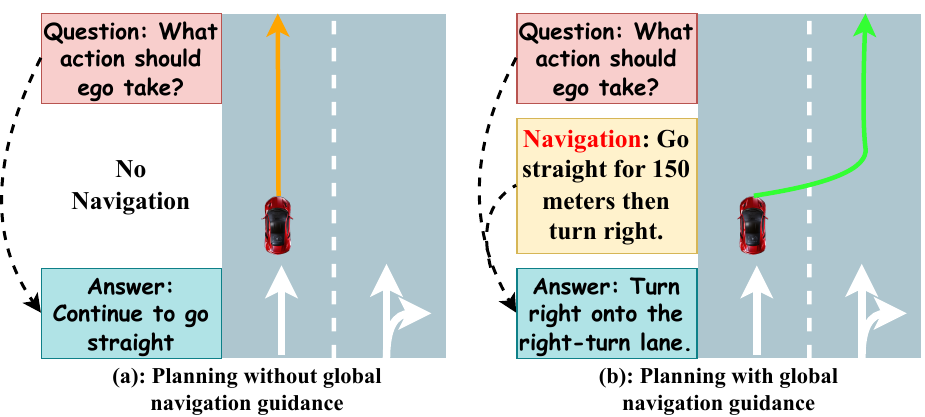}
    \caption{Comparison between a) planning without global navigation guidance and b) planning with global navigation guidance. In this example, the vehicle needs to turn right at the next corner. Without beyond-view-range (BVR) knowledge from navigation, the planner makes a conservative decision to continue straight. With global BVR knowledge, it appropriately directs the vehicle to merge into the right-turn lane. Concrete examples from experiments are shown in Fig. \ref{fig:open-vis} and Fig. \ref{fig:close-vis}.  }
    \label{fig:teaser}
\end{figure}

Autonomous driving systems \cite{chen2024end,chib2023recent,24mm01,24mm02,24mm03, jiabench2drive,nie2024reason2drive} have made remarkable progress in recent years, enabling vehicles to perceive their immediate surroundings, predict the movement of nearby objects, and plan appropriate actions. These systems can be categorized into two main types: vision-language models (VLMs) \cite{sima2024drivelm,xu2024drivegpt4,ding2024nuinstruct,marcu2024lingoqa} for question-answering tasks and end-to-end driving models \cite{jiang2023vad,chen2024vadv2,sun2024sparsedrive,zheng2024genad} for perception, prediction, and planning. However, these approaches primarily rely on responses to agent or environment within visual range (normally within 100 and 150 meters in autonomous vehicles), which creates a critical gap in incorporating global contextual information for \emph{human-like} long-term planning. This limitation constrains both VLMs and end-to-end models, hindering their ability to reason proactively and generalize to unfamiliar scenes.

In real-world driving scenarios, navigation applications such as Google Maps \cite{svennerberg2010api} serve as essential tools that provide global contextual information for human drivers. These applications communicate the ego vehicle's intended future maneuvers (e.g., turning left or right, proceeding straight) alongside three critical pieces of information: distance to upcoming maneuvers, intersection type, and presence of traffic signals. \emph{Notably, the distance information typically extends beyond the visual perception capabilities of onboard sensors such as cameras or LiDAR, and is therefore classified as beyond visual range (\textbf{BVR}) \cite{merz2011bvr1,dantas2023bvr2} information.} Despite being crucial for effective planning and decision-making, BVR information remains largely unexplored in autonomous driving research. Current Q\&A datasets \cite{sima2024drivelm,ding2024nuinstruct} and models \cite{jiang2023vad,sun2024sparsedrive} predominantly focus on frame-by-frame perception and prediction, or at most extend to the next frame for supervision \cite{li2025navigationguided}, without adequately addressing the global navigation context necessary for comprehensive scene understanding and long-term planning.

In Fig. \ref{fig:teaser}, we demonstrate how navigation guidance enhances both question-answering performance and end-to-end planning. In this scenario, navigation provides critical information indicating an intersection located 150 meters ahead where the ego vehicle must execute a right turn. However, due to the limited perception range of onboard sensors—typically restricted to between 100 and 150 meters—the ego vehicle cannot detect this intersection with sufficient advance notice to initiate the necessary lane change for the right-turn lane. In contrast, by incorporating global BVR knowledge from navigation tools, the planner proactively directs the vehicle to merge into the right-turn lane well in advance of the upcoming turn, demonstrating the tangible benefits of navigation-guided planning.


To address this gap, we propose NavigScene, an auxiliary navigation-based dataset derived from the nuScenes \cite{caesar2020nuscenes} and NAVSIM \cite{dauner2024navsim} datasets. Through natural language navigation instructions, we simulate a \emph{human-like} driving environment within autonomous driving systems, effectively imitating navigation tools such as Google Maps that provide BVR knowledge critical for driving decisions and planning. NavigScene comprises two subsets: NavigScene-nuScenes and NavigScene-NAVSIM. Our dataset bridges the disconnection between local sensor data and global navigation context by providing paired data: multi-view sensor inputs (images or videos) alongside corresponding natural language navigation guidance that captures the global driving environment. This carefully constructed pairing enables autonomous systems to reason more comprehensively about driving scenarios and make more informed planning decisions that emulate human behaviors guided by navigation applications.

Building upon the \emph{human-mimicking} auxiliary dataset NavigScene, we propose three paradigms to leverage navigation guidance in autonomous driving tasks like Q\&A, perception, prediction and planning. First is navigation-guided reasoning, which can be implemented through Navigation-guided Supervised Fine-tuning (\textbf{NSFT}) for driving-related Q\&A tasks. By incorporating navigation guidance into prompts, we enable more comprehensive reasoning that considers both local visual cues and global navigational context, significantly improving the model's ability to answer questions requiring knowledge beyond the immediate visual range. Second is a reinforcement learning method named Navigation-guided Preference Optimization (\textbf{NPO}). We introduce an auxiliary text summarization task to enhance Direct Preference Optimization (DPO) \cite{rafailov2023dpo} by establishing a preference relationship between summarized answers from the vision-language model and navigation guidance, thereby improving vision-language models' BVR reasoning and generalization capabilities. \emph{The post-training of VLMs with navigation guidance consists of NSFT and NPO.} Third is the Navigation-guided Vision-Language-Action (\textbf{NVLA}) model. We develop a VLA baseline architecture that integrates navigation guidance and vision-language models with conventional end-to-end driving models through feature fusion, creating a more robust representation for downstream tasks including perception, prediction, and planning. Our contributions can be summarized in three main aspects:

\setlist{nolistsep}
\begin{itemize}[noitemsep,leftmargin=*]
    \item We propose NavigScene, a novel auxiliary dataset that pairs local multi-view sensor inputs with global natural language navigation guidance, addressing the critical gap between local perception and global navigation context in autonomous driving.
    
    \item We implement NavigScene across three complementary paradigms: navigation-guided reasoning, navigation-guided preference optimization, and a navigation-guided vision-language-action model, enhancing autonomous driving systems' reasoning and generalization capabilities beyond visual range limitations.
    
    \item We conduct comprehensive experiments on both Q\&A tasks and end-to-end driving tasks—including perception, prediction, and planning—demonstrating the significant performance improvements achieved by incorporating global navigation knowledge into autonomous driving systems.
\end{itemize}



\section{Related Works}

\textbf{LLMs and VLMs in Autonomous Driving. } NuScenes-QA \cite{qian2024nuscenes-qa} is the first Visual Question Answering benchmark specifically designed for autonomous driving scenarios, establishing a foundation with several baselines that leverage advanced 3D detection and VQA techniques. DriveGPT4 \cite{xu2024drivegpt4} introduces an interpretable end-to-end autonomous driving system powered by Large Language Models, while DriveLM \cite{sima2024drivelm} enhances Visual Language Models' reasoning capabilities through graph-based visual question answering. NuInstruct \cite{ding2024nuinstruct} places greater emphasis on crucial multi-view and temporal information in Visual Language Models, which is essential for robust autonomous driving systems. VLP \cite{pan2024vlp} proposes a novel framework that exploits LLMs to bridge the gap between linguistic understanding and
autonomous driving. 

\noindent\textbf{End-to-end Autonomous Driving. }VAD \cite{jiang2023vad} employs an ego query mechanism to predict single-mode trajectories, while VADv2 \cite{chen2024vadv2} advances this approach by implementing a probabilistic space based on multiple trajectories. SparseDrive \cite{sun2024sparsedrive} innovates by designing parallel motion and planning modules that reduce computational demands from BEV features. DiffusionDrive \cite{diffusiondrive} introduces a truncated diffusion policy to enhance the trajectory's probabilistic representation, while MomAD \cite{song2025momad} focuses on improving stability and maintaining consistency across consecutive planning decisions.

\section{Navigation-based Datasets: NavigScene}


\begin{figure*}[!ht]
    \centering
    \includegraphics[width=1.0\linewidth]{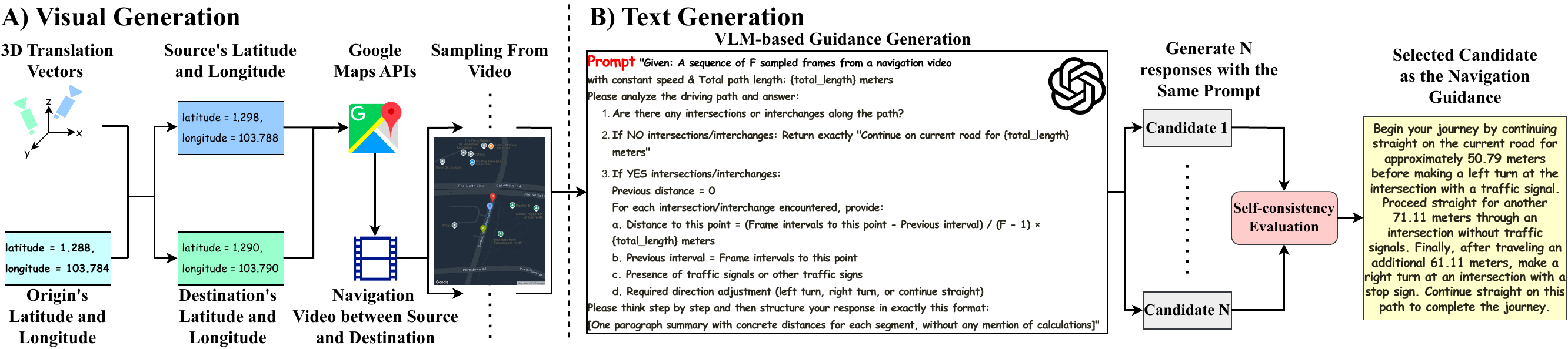}
    \caption{Navigation guidance generation process of one scene. Part A (Visual Generation): Source and destination coordinates are calculated using the origin's coordinate and 3D translation vectors. A navigation video is constructed via Google Maps APIs, then evenly sampled to extract multiple frames. Part B (Text Generation): The multiple frames are processed by a vision-language model (GPT-4o \cite{achiam2023gpt4o}) with a specialized prompt to generate several candidate responses. Self-consistency evaluation selects the highest-scoring candidate as the final navigation guidance.}
    \label{fig:dataset}
\end{figure*}

Existing autonomous driving datasets predominantly emphasize local-level descriptions of key frames, which serve perception tasks effectively. However, they do not adequately address the beyond-visual-range (BVR) knowledge essential for scene understanding and decision making. This limitation is particularly significant given that navigation applications are integral components of real-world driving scenarios, especially when experienced human drivers are traversing unfamiliar territories or assessing current road conditions. To simulate \emph{human-like} driving environments and bridge this gap, we propose NavigScene, an auxiliary navigation-guided natural language dataset that simulates navigation applications to provide global contextual information, thereby enhancing the capability of autonomous systems to reason and generalize with BVR knowledge in complex driving environments.

Our proposed NavigScene dataset is derived from scenes in the nuScenes \cite{caesar2020nuscenes} dataset and NAVSIM \cite{dauner2024navsim}, and consists of two subsets: NavigScene-nuScenes and NavigScene-NAVSIM. The generation process for a single scene is shown in Fig. \ref{fig:dataset}.

\subsection{Visual Generation}
\label{sec:vis-gen}

To establish each scene, we first determine the latitudes and longitudes of both the source and destination. This calculation incorporates the origin's coordinates and the 3D translation vectors of the source and destination from this origin. For an origin with coordinates $(\phi, \lambda)$, where $\phi$ represents latitude and $\lambda$ represents longitude (both in decimal degrees), and a translation vector $(\Delta x, \Delta y, \Delta z)$ in meters (where $\Delta x$ denotes the eastward component, $\Delta y$ the northward component, and $\Delta z$ the upward component), the coordinates of the source or destination $(\phi', \lambda')$ can be calculated using \cite{strang1997longitude}:
\begin{equation}
\label{eq:latitude}
    \phi' = \phi + \frac{180}{\pi} \cdot \frac{\Delta y}{R}, \quad \quad \lambda' = \lambda + \frac{180}{\pi} \cdot \frac{\Delta x}{R \cdot \cos(\frac{\pi}{180} \cdot \phi)},
\end{equation} where $R$ represents the Earth's radius, approximately 6,378,137 m. While $\Delta x$ and $\Delta y$ represent translations in the eastern and northern directions in meters, $\Delta z$ is excluded from latitude and longitude calculations as it does not influence horizontal positioning.

We leverage Google Maps APIs \cite{svennerberg2010api} to generate navigation videos using these coordinates. The Direction API provides precise routes, the Static Map API acquires sequential images along routes, and the Distance Matrix API estimates driving distance and duration. Assuming constant velocity, we synthesize realistic navigation videos simulating the driving experience. To facilitate analysis, we evenly sample $F$ frames from these videos for subsequent text generation via VLM in Sec. \ref{sec:text-gen}.

\subsection{Text Generation}
\label{sec:text-gen}

While Sec. \ref{sec:vis-gen} on visual generation (Part A in Fig. \ref{fig:dataset}), integrating complete navigation videos with VLMs or end-to-end architectures poses significant challenges due to alignment difficulties between navigation videos and sensor-based training data. To address this limitation, we transform sequential images into natural language navigation descriptions using VLMs (Part B in Fig. \ref{fig:dataset}). For each sequence of frames, we generate $N$ candidate navigations through a specialized prompt shown in Fig. \ref{fig:dataset}. This prompt first analyzes intersections or interchanges to determine driving directions, then estimates distances based on frame intervals.

After obtaining $N$ candidate responses, we implement a novel selection strategy to identify the optimal description. We define three similarity metrics $S_{inter}(\cdot, \cdot)$, $S_{dist}(\cdot, \cdot)$, and $S_{word}(\cdot, \cdot)$:

$S_{inter}(\cdot, \cdot)$ represents intersection similarity, emphasizing directional keywords accuracy. For candidate $a_i$, directional keywords are extracted as $K_{inter}(a_i) = (m_1^i, m_2^i, \ldots)$, then the intersection similarity between $a_i$ and $a_j$ is:
\begin{equation}
S_{inter}(a_i, a_j) = \begin{cases}
1, & \text{if } |K_{inter}(a_i)| = |K_{inter}(a_j)| \\
& \text{ and } m_d^i = m_d^j \text{ for all } d \\
0, & \text{otherwise}
\end{cases}
\end{equation}

$S_{dist}(\cdot, \cdot)$ represents distance value similarity. Distance values in $a_i$ are $K_{dist}(a_i) = (n_1^i, n_2^i, \ldots)$, then the distance similarity is:
\begin{equation}
S_{dist}(a_i, a_j) = \begin{cases}
\mathbb{E}_{1 \leq d \leq |K_{dist}(a_i)|}\left[1 -\frac{|n_d^i - n_d^j|}{max(n_d^i, n_d^j)}\right], \text{otherwise } \\
0,  \text{if } |K_{dist}(a_i)| = |K_{dist}(a_j)|
\end{cases}
\end{equation}

$S_{word}(\cdot, \cdot)$ represents lexical similarity, calculated using the Jaccard index:
\begin{equation}
S_{word}(a_i, a_j) = |a_i \cap a_j| / |a_i \cup a_j|.
\end{equation}

The overall similarity score $S_{over}(\cdot, \cdot)$ between candidates is:
\begin{equation}
S_{over}(a_i, a_j) = \eta_1 S_{inter}(a_i, a_j) + \eta_2 S_{dist}(a_i, a_j) + \eta_3 S_{word}(a_i, a_j)
\end{equation}

Given that directional accuracy is most critical, followed by distance precision and then lexical similarity, we assign weights such that $\eta_1 > \eta_2 > \eta_3$, then select the optimal answer $a^*$ by identifying the candidate with the highest cumulative similarity:
\begin{equation}
a^* = \arg\max_{a_i \in A} \sum_{j \neq i} S_{over}(a_i, a_j)
\end{equation}

This approach identifies the best candidate, which serves as the final navigation to simulate \emph{human-like} driving environment.

\section{Methodology}

\subsection{Navigation-guided Reasoning}
\label{sec:reason}

\begin{figure}[!ht]
    \centering
    \includegraphics[width=1.0\linewidth]{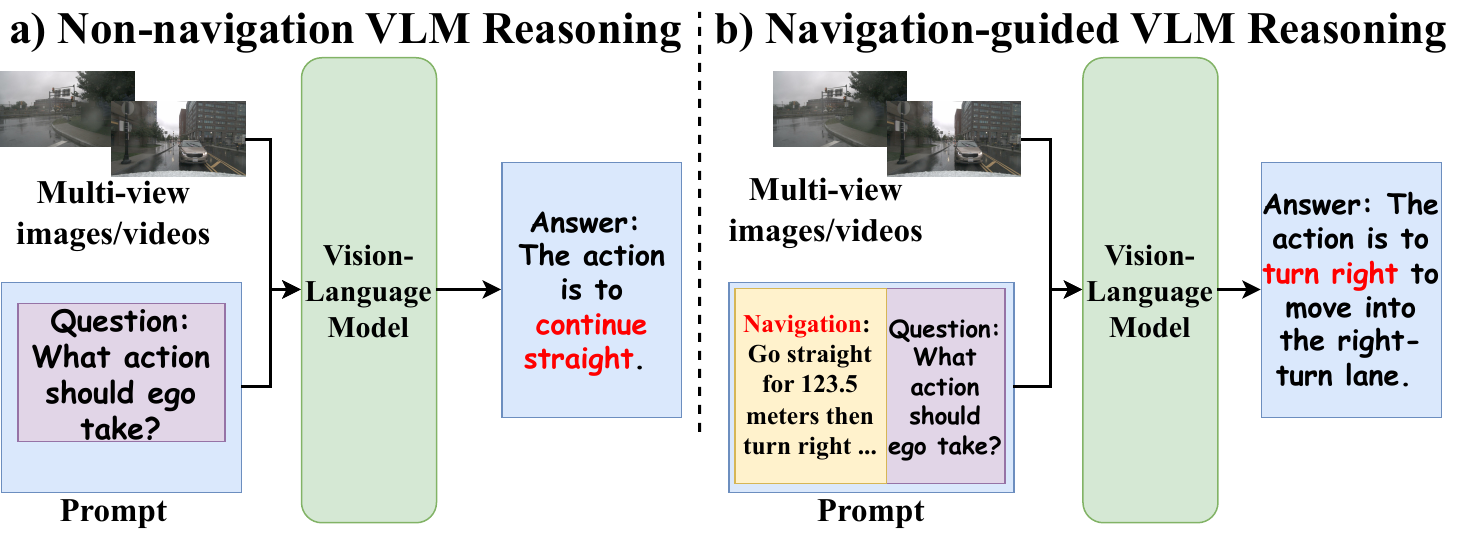}
    \caption{Comparison between a) non-navigation VLM reasoning and b) navigation-guided VLM reasoning. In our proposed navigation-guided paradigm, both the navigation guidance and question together form the prompt for VLM. (Best viewed when zoomed in.)}
    \label{fig:reason}
\end{figure}

In traditional Q\&A tasks for autonomous driving \cite{sima2024drivelm,ding2024nuinstruct}, multi-view images or videos paired with questions serve as input for VLM, as shown in Fig. \ref{fig:reason}a. However, this reasoning paradigm is limited to in-scope information and overlooks beyond-visual-range (BVR) information, which is crucial for long-term planning. To address this limitation, we incorporate navigation guidance into our prompting approach (as shown in Fig. \ref{fig:reason}b), thereby enriching the reasoning process with essential global information. 

The navigation-guided reasoning can be represented as:
\begin{equation}
\label{eq:reason}
a = M(v, g \oplus q),
\end{equation}
where $M$ is the vision-language model, $a$ represents natural language outputs as the answer, $v$ denotes multi-view images or videos, $g$ is the navigation guidance, and $q$ is the question. The concatenation of $g$ and $q$ serves as the prompt for $M$. \emph{This paradigm is applied in the Navigation-guided Supervised Fine-Tuning (\textbf{NSFT}) of driving-related Q\&A task in this paper.}

\subsection{Navigation-guided Preference Optimization} 
\label{sec:npo}
While Section \ref{sec:reason} introduced NSFT to enhance VLM reasoning capabilities, this approach has limitations in generalizing to unseen scenarios. For VLMs with fewer than 10B parameters, supervised fine-tuning restricts generalization ability \cite{dong2024abilities}, and reasoning capacity is constrained by parameter scale \cite{bi2025enhancing}. To address these shortcomings, we propose Navigation-guided Preference Optimization (\textbf{NPO}), an extension of Direct Preference Pptimization (DPO) \cite{rafailov2023dpo} applied after NSFT to improve generalization performance on novel navigation scenarios.

NPO builds upon DPO by integrating navigation-related knowl-
edge through an auxiliary text summarization task. In NPO, the VLM still processes multi-view images $v$ and question $q$ as inputs. We establish a preference relationship between detailed answer $a$ and its summarized version $s$. The summarization $s$ is generated online using a VLM $M$:
\begin{equation}
s = M(\text{pt}\oplus a),
\end{equation}
where $a$ is the original answer from supervised fine-tuning, and pt is the prompt \emph{"Summarize this answer to a driving-relevant question to make it simple without losing important information."} 

Following DPO methodology, we initialize a learnable reward model $M^{\theta}$ and a frozen reference model $M^*$. Thus, we can obtain $s^\theta$ and $s^*$ from these two VLMs respectively. With navigation guidance $g$, we quantify the quality of the summarized answer $s$ using the mutual information \cite{tishby2015ib}:
\begin{equation} \label{eq:mi}
\begin{aligned}
\text{mi}(s) &= p(a, s)\log \frac{p(a,s)}{p(a)p(s)} - p(s, g)\log \frac{p(s,g)}{p(s)p(g)},\\
&= -\log p(s) - p(s)p(g|s) \log \frac{p(g|s)}{p(g)}.
\end{aligned}
\end{equation}

Eq. \ref{eq:mi} has two goals: to simplify the summarized answer $s$ compared to the original answer $a$, while enhancing the relevance between the summarized answer $s$ and the navigation guidance $g$. Based on the implementation in \cite{west2019bottlesum}, Equation \ref{eq:mi} is further simplified to:
\begin{equation} \label{eq:mi-simple}
\text{mi}(s) = -\log p(s) - p(s) \log p(g|s).
\end{equation}

By incorporating this measurement, we define the reward for summarized answers as:
\begin{equation} \label{eq:sum-reward}
r_s = \log p^{\theta}(s^{\theta}|v, q) - \log p^{*}(s^{*}|v, q) + \alpha[\text{mi}(s^\theta) - \text{mi}(s^*)],
\end{equation} 
where $\alpha$ is a trade-off hyper parameter. This reward not only measures the difference in summarized answer $s$ between reward model $M^\theta$ and reference model $M^*$, but also the difference in guidance relevance between the two summarized answers.

Similarly, the reward for the original answer $a$ is:
\begin{equation} \label{eq:ans-reward}
r_a = \log p^{\theta}(a|v, q) - \log p^{*}(a|v, q).
\end{equation}

The objective function of NPO is thus formulated as:
\begin{equation}
\mathcal{L}_{\text{NPO}}(\theta) = - \mathbb{E}_{(v,q,a,s^{\theta},s^{*})\in D}[\log \sigma(r_s - r_a)]
\end{equation}
where $\sigma$ is the sigmoid function, and $D$ represents the preference dataset that contains tuples of (multi-view images, question, answer, summary from reward model, summary from reference model).

In our proposed NPO method, we introduce an auxiliary task, navigation-guided text summarization, for both the reward model and the reference model. This strategic addition directs the reward model to focus on guidance-relevant knowledge, significantly enhancing its ability to generate driving-relevant, concise responses while preserving critical information aligned with navigation requirements. When integrated with the end-to-end driving model in Sec. \ref{sec:vla}, this paradigm substantially improves the overall autonomous driving system's generalization capability.

\subsection{Navigation-guided Vision-Language-Action Model}
\label{sec:vla}

\begin{figure}[!ht]
    \centering
    \includegraphics[width=1.0\linewidth]{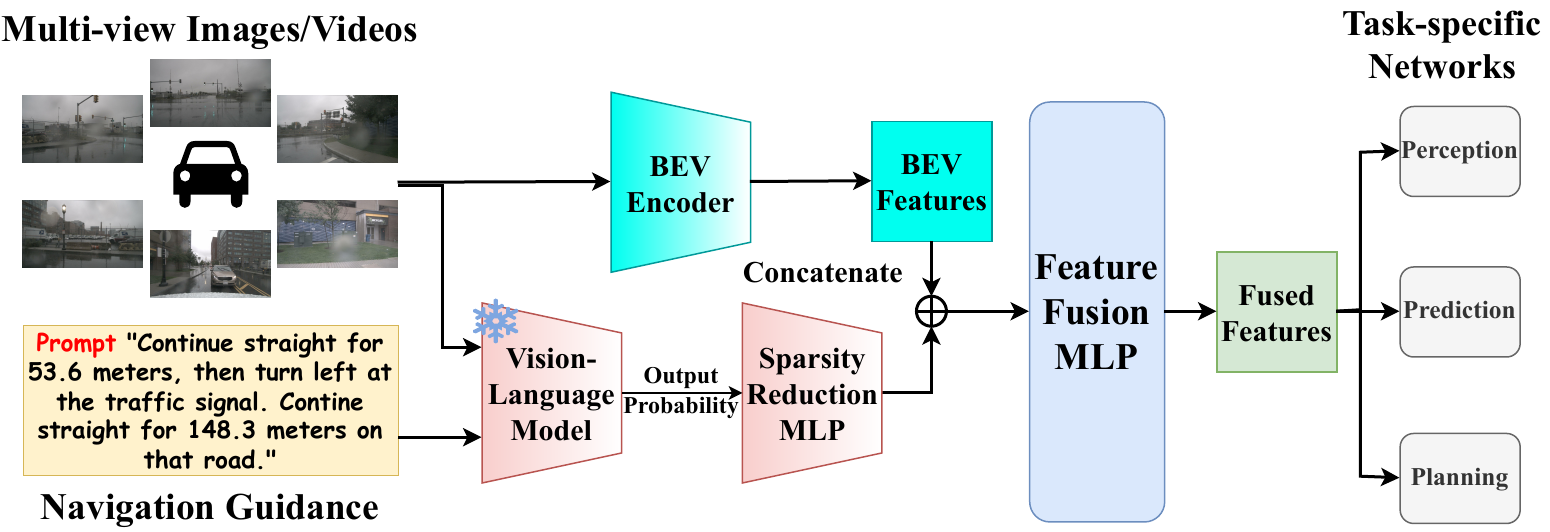}
    \caption{Navigation-guided vision-language-action model for end-to-end driving. BEV features are concatenated with vision-language features generated by the frozen VLM and a learnable sparsity reduction MLP, then processed through a learnable feature fusion MLP to produce fused features for task-specific networks.}
    \label{fig:vla}
\end{figure}

Beyond question-answering tasks, navigation guidance substantially enhances end-to-end driving system's performance. Conventional end-to-end models \cite{jiang2023vad,sun2024sparsedrive} that rely solely on sensor data (multi-view images or videos) suffer from limited reasoning capabilities and poor generalization to novel scenarios. To address these shortcomings, we propose a Navigation-guided Vision-Language-Action (\textbf{NVLA}) model that integrates navigation guidance with vision-language models into the end-to-end driving framework.

In conventional end-to-end models, the output of a driving task $j \in \{\text{perception}, \text{prediction}, \text{planning}\}$ can be represented as:
\begin{equation}
\label{eq:e2e_old}
    o_{j}^{\text{con}} = H_{j}(E(v)),
\end{equation}
where $v$ denotes multi-view images or videos, $E$ is the BEV encoder, $H_{j}$ is the task-specific network, and $o_{j}^{\text{con}}$ is the output of task $j$ without navigation guidance.

In our navigation-guided VLA model, we incorporate both a VLM post-trained by NSFT and NPO, and navigation guidance. The output probability distribution of modern VLMs typically has a high dimensionality due to their large vocabulary space—for example, LlamaAdapter's \cite{zhang2024llamaadapter} output probability dimension is 32,000—making direct alignment or fusion with BEV features (typically 256 dimensions in models like SparseDrive \cite{sun2024sparsedrive}) challenging. To address this mismatch, we introduce a learnable sparsity reduction MLP $\phi^{\text{red}}$ to compress VLM features' dimension, followed by a learnable feature fusion MLP $\phi^{\text{fus}}$. The complete process is represented as:
\begin{equation}
\label{eq:e2e_new}
    o_{j}^{\text{nav}} = H_{j}(\phi^{\text{fus}}(E(v) \oplus \phi^{\text{red}}(M(v, g)))),
\end{equation}
where $M$ represents the frozen VLM that has been trained via NSFT as described in Sec. \ref{sec:reason} then NPO as described in Sec. \ref{sec:npo}, $g$ is the navigation guidance, and $o_{j}^{\text{nav}}$ is the output of task $j$ with navigation guidance. 
This integration process is illustrated in Fig. \ref{fig:vla}.

\section{Experiments}

\vspace{-8pt}
\begin{table}[!ht]
    \scriptsize
    \centering
    \caption{Reasoning Results on DriveLM-nuScenes}
    \resizebox{1.0\linewidth}{!}{%
    \begin{tabular}{c|c|ccccccc}
          \toprule
          VLM & NavigScene &BLEU-4 $\uparrow$ & METEOR $\uparrow$ & CIDEr $\uparrow$ & ROUGE\_L $\uparrow$ & SPICE $\uparrow$ & GPT $\uparrow$ & Comp. $\uparrow$ \\
         \hline
         \multirow{2}{*}{Llama-Adapter} & $\times$ & 50.68 & 33.75  & 2.37 & 64.59 & 44.20 & 70.83 & 29.98\\
          & $\checkmark$ & 54.25 & 37.62 & 2.81 & 67.66 & 48.35 & 74.08 & 33.07 \\
         \hline
         \multirow{2}{*}{Llava-7B} & $\times$ & 49.75 & 33.21  & 2.19 & 63.84 & 42.56 & 70.24 & 29.37 \\
         & $\checkmark$ & 53.93 & 36.86 & 2.75 & 66.97 & 46.83 & 73.77 & 32.79\\
         \hline
         \multirow{2}{*}{Qwen2.5-7B} & $\times$ & 51.65 & 34.12  & 2.46 & 64.97 & 46.45 & 71.29 & 30.31\\
         & $\checkmark$ & 55.13 & 38.20 & 3.14 & 67.88 & 49.89 & 74.87 & 34.26\\    
         \bottomrule
    \end{tabular}%
    }
\label{tab:drivelm}
\end{table}

\begin{table*}[!ht]
    \scriptsize
    \centering
    \caption{Reasoning Results on NuInstruct}
    \resizebox{1.0\linewidth}{!}{%
    \begin{tabular}{c|c|cccccc|cc|cccccc|c}
          \toprule
          \multirow{2}{*}{VLM} & \multirow{2}{*}{NavigScene} & \multicolumn{6}{c|}{\textbf{Perception}} & \multicolumn{2}{c|}{\textbf{Prediction}} & \multicolumn{6}{c|}{\textbf{Risk}} & \multirow{2}{*}{\textbf{Planning} $\uparrow$} \\
          & & Dis $\downarrow$ & Spe $\downarrow$ & Ins $\downarrow$ & Clo $\uparrow$ & Sta $\uparrow$ & SaR $\uparrow$ & Mot $\downarrow$ & Sta $\uparrow$ & App $\uparrow$ & Lan $\uparrow$ & Onc $\uparrow$ & Cro $\uparrow$ & Ove $\uparrow$ & Bra $\uparrow$ & \\
         \hline
         \multirow{2}{*}{Llama-Adapter} & $\times$ & 27.9 & 6.6 & 6.5 & 20.4 & 16.2 & 24.0 & 8.7 & 40.2 & 13.5 & 18.7 & 16.2 & 18.4 & 6.5 & 21.5 & 25.7 \\
          & $\checkmark$ & 24.3 & 3.9 & 4.0 & 32.2 & 19.8 & 28.7 & 4.3 & 44.1 & 15.8 & 21.0 & 19.9 & 22.8 & 9.0 & 26.4 & 31.2 \\
         \hline
         \multirow{2}{*}{Llava-7B} & $\times$ & 28.4 & 6.9 & 6.6 & 22.1 & 16.2 & 23.5 & 9.4 & 38.9 & 12.3 & 19.6 & 16.9 & 18.7 & 6.5 & 21.3 & 25.3 \\
          & $\checkmark$ & 24.5 & 4.3 & 4.1 & 27.6 & 19.5 & 27.8 & 6.3 & 43.6 & 15.0 & 22.7 & 20.2 & 23.0 & 9.2 & 26.8 & 32.6 \\
         \hline
         \multirow{2}{*}{Qwen2.5-7B} & $\times$ & 26.9 & 5.6 & 5.4 & 26.5 & 17.3 & 26.7 & 7.8 & 41.8 & 15.2 & 20.3 & 18.4 & 20.5 & 7.7 & 22.1 & 26.6 \\
          & $\checkmark$ & 23.6 & 3.1 & 3.2 & 33.7 & 20.2 & 31.9 & 3.8 & 45.3 & 18.6 & 23.7 & 21.9 & 26.2 & 10.4 & 27.9 & 36.4 \\    
         \bottomrule
    \end{tabular}%
    }
\label{tab:NuInstruct}
\end{table*}

\noindent\textbf{Datasets.} In this paper, we evaluate our proposed NavigScene alongside two categories of benchmark datasets. The first category comprises Q\&A datasets designed for supervised fine-tuning and reinforcement learning of VLMs. The second category includes end-to-end driving datasets that assess the integration of NavigScene and navigation-guided VLMs with autonomous driving models.

For Q\&A datasets, we utilize DriveLM-nuScenes \cite{sima2024drivelm}, which contains approximately 700 scenes, each with several hundred questions corresponding to images from six camera views. In our experiments, we allocate 200 scenes for testing and the remainder for training. We also employ NuInstruct \cite{ding2024nuinstruct}, another Q\&A dataset containing over 10,000 pairs with images from six camera perspectives. Since both datasets are derived from nuScenes \cite{caesar2020nuscenes}, it is methodologically sound to incorporate NavigScene-nuScenes in their training processes.

For end-to-end driving evaluation, we employ two benchmark datasets: the open-loop nuScenes dataset \cite{caesar2020nuscenes} paired with NavigScene-nuScenes, and the closed-loop NAVSIM dataset \cite{dauner2024navsim} paired with NavigScene-NAVSIM.  nuScenes features 1,000 driving scenes captured in Boston and Singapore, each spanning 20 seconds with comprehensive 3D bounding box annotations. It includes synchronized data from six cameras, one LiDAR sensor, and five radars. NAVSIM is a simulation dataset based on nuPlan \cite{caesar2021nuplan}, comprising 120 hours of driving data from eight cameras and five LiDAR sensors.

\noindent\textbf{Implementations.} For NavigScene generation, we set $F=20$, $N=5$, with weights $\eta_1=0.5$, $\eta_2=0.3$, and $\eta_3=0.2$. For VLMs, we evaluate Llama-Adapter-7B \cite{zhang2024llamaadapter}, Llava-v1.6-Mistral-7B \cite{liu2023llava}, and Qwen2.5-VL-Instruct-7B \cite{yang2024qwen25}. During fine-tuning and preference optimization, we apply LoRA with rank 16, learning rate 1e-4, and $\alpha=0.6$. The output sequence length is limited to 128 tokens while maintaining default values for other parameters. For end-to-end driving evaluation, we utilize VAD \cite{jiang2023vad} and SparseDrive \cite{sun2024sparsedrive}. The sparsity reduction MLP's input channels match each VLM's vocabulary dimension, while its output channels align with the original BEV features. The feature fusion MLP receives inputs with twice the original BEV feature dimension and outputs tensors matching the original BEV features. All MLPs use a learning rate of 2e-4, while preserving default values for other network parameters. All models employ AdamW \cite{loshchilovdecoupled} as the optimizer. For NSFT and NVLA, epochs are assigned with the same with the original training of VLMs and end-to-end models, while NPO are conducted for 10 epochs. All experiments are conducted three times on Nvidia H800 GPUs.

\begin{figure*}[!ht]
    \centering
    \includegraphics[width=1.0\linewidth]{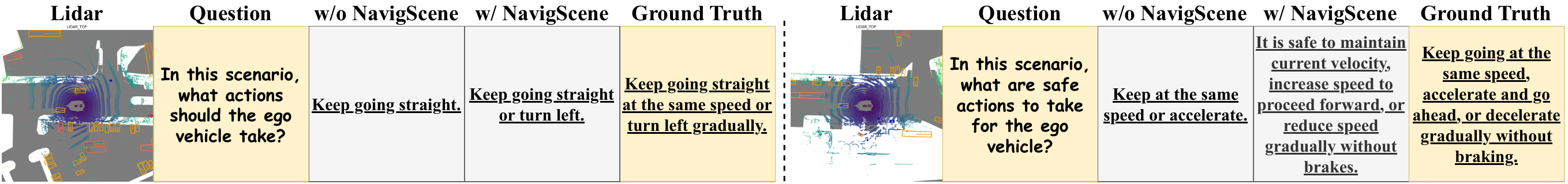}
    \caption{Examples of question-answering on the DriveLM dataset. }
    \label{fig:drivelm-vis}
\end{figure*}

\begin{figure*}[!ht]
    \centering
    \includegraphics[width=1.0\linewidth]{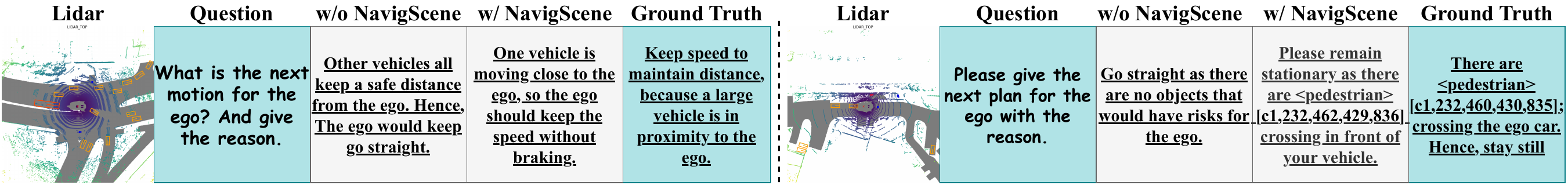}
    \caption{Examples of question-answering on the NuInstruct dataset. }
    \label{fig:nuinstruct-vis}
\end{figure*}

\subsection{Quantitative Results on Q\&A Tasks}

In Tab. \ref{tab:drivelm}, we compare the performance of VLMs post-trained with NavigScene via NSFT and NPO against baselines only using DriveLM \cite{sima2024drivelm} across three open-source VLMs. Following DriveLM's evaluation protocol, we employ metrics including BLEU-4 \cite{papineni2002bleu}, METEOR \cite{banerjee2005meteor}, CIDEr \cite{vedantam2015cider}, ROUGE\_L \cite{lin2004rouge}, and SPICE \cite{anderson2016spice}, along with GPT score \cite{achiam2023gpt4o,sima2024drivelm} and completeness (Comp.) \cite{sima2024drivelm}. Results consistently demonstrate that NavigScene significantly enhances the quality of driving-relevant responses across all these VLMs.

In Tab. \ref{tab:NuInstruct}, we present results from NuInstruct \cite{ding2024nuinstruct} comparing baseline VLMs with models post-trained with NavigScene. Following the NuInstruct evaluation protocols, we assess perception capabilities through subtasks measuring distance (Dis), speed (Spe), instance number (Ins), closest object identification (Clo), status recognition (Sta), and same-road detection (SaR). For prediction, we evaluate motion trajectory (Mot) and future status (Sta) forecasting. Risk assessment is measured across six dimensions: approach (App), lane changing (Lan), ongoing maneuvers (Onc), crossing (Cro), overtaking (Ove), and braking (Bra). Results illustrate that NavigScene significantly improves performance across all tested VLMs, enhancing their reasoning capabilities in driving-related Q\&A tasks.

\subsection{Qualitative Results on Q\&A Tasks}

In Fig. \ref{fig:drivelm-vis} and Fig. \ref{fig:nuinstruct-vis}, we show examples of VLM responses both with and without NavigScene integration. The BVR knowledge provided by NavigScene significantly enhances the VLM's reasoning capabilities, resulting in more complete and accurate answers. Additional samples are exhibited in the \textcolor{blue}{Supplementary Material} (\textcolor{blue}{SM}).

\subsection{Quantitative Results on End-to-end Driving}

\begin{table*}[!ht]
    \scriptsize
    \centering
    \caption{Results for both open-loop and closed-loop planning settings. Left: Open-loop planning performance on nuScenes. Right: Closed-loop planning performance on NAVSIM. All VLMs underwent post-training via NSFT then NPO on \textcolor{blue}{DriveLM-nuScenes} and NavigScene-nuScenes, followed by prompting with NavigScene-nuScenes and NavigScene-NAVSIM respectively during training.}
    \resizebox{1.0\linewidth}{!}{%
    \begin{tabular}{c|c|cccc|cccc|ccccc|c}
          \toprule
          \multirow{3}{*}{\begin{tabular}[c]{@{}c@{}}End-to-end\\Model\end{tabular}} & \multirow{3}{*}{VLM} & \multicolumn{8}{c|}{\textbf{Open-loop Planning on nuScenes}} & \multicolumn{6}{c}{\textbf{Closed-loop Planning on NAVSIM}} \\
          \cline{3-16}
          & & \multicolumn{4}{c|}{\textbf{L2 (m)$\downarrow$}} & \multicolumn{4}{c|}{\textbf{Collision Rate (\textpertenthousand) $\downarrow$}} & \multirow{2}{*}{NC $\uparrow$} & \multirow{2}{*}{DAC $\uparrow$} & \multirow{2}{*}{TTC $\uparrow$} & \multirow{2}{*}{Comf. $\uparrow$} & \multirow{2}{*}{EP $\uparrow$} & \multirow{2}{*}{PDMS $\uparrow$} \\
          \cline{3-10}
          & & 1s & 2s & 3s & Avg. & 1s & 2s & 3s & Avg. & & & & & & \\
         \hline
         \multirow{4}{*}{VAD} & None & 0.54 & 1.15 & 1.98 & 1.22 & 9.76 & 24.20 & 95.93 & 43.30 & 97.0 & 86.5 & 89.7 & 100 & 75.4 & 80.6 \\
         & Llama-Adapter & 0.42 & 1.02 & 1.76 & 1.07 & 6.28 & 21.85 & 81.37 & 36.50 & 97.9 & 88.8 & 91.6 & 100 & 77.6 & 82.2 \\
         & Llava-7B & 0.45 & 1.04 & 1.79 & 1.09 & 6.93 & 22.09 & 83.41 & 37.48 & 97.7 & 88.1 & 91.3 & 100 & 76.8 & 81.8 \\
         & Qwen2.5-7B & 0.37 & 0.96 & 1.65 & 0.99 & 4.52 & 18.10 & 74.62 & 32.41 & 98.3 & 90.6 & 92.4 & 100 & 79.0 & 84.0 \\
         \hline
         \multirow{4}{*}{\begin{tabular}[c]{@{}c@{}}Sparse\\Drive\end{tabular}} & None & 0.44 & 0.92 & 1.69 & 1.01 & 7.38 & 19.46 & 70.60 & 32.48 & 97.2 & 91.7 & 91.4 & 100 & 77.9 & 82.4 \\
         & Llama-Adapter & 0.33 & 0.78 & 1.47 & 0.86 & 5.97 & 15.65 & 58.41 & 26.68 & 98.0 & 93.5 & 93.0 & 100 & 80.4 & 83.9 \\
         & Llava-7B & 0.36 & 0.80 & 1.46 & 0.87 & 6.26 & 16.34 & 62.58 & 28.39 & 98.1 & 93.3 & 92.8 & 100 & 78.6 & 83.1 \\
         & Qwen2.5-7B & 0.29 & 0.64 & 1.35 & 0.76 & 4.38 & 12.99 & 44.75 & 20.71 & 98.3 & 96.0 & 94.1 & 100 & 81.7 & 86.5 \\
         \bottomrule
    \end{tabular}%
    }
\label{tab:planning}
\end{table*}

\begin{table*}[!ht]
    \scriptsize
    \centering
    \caption{Object detection, tracking, mapping, and motion forecasting on nuScenes. All VLMs were first post-trained via NSFT then NPO on \textcolor{blue}{DriveLM-nuScenes} and NavigScene-nuScenes, then prompted using NavigScene-nuScenes during training.}
    \resizebox{1.0\linewidth}{!}{%
    \begin{tabular}{c|c|cccc|cccc|cccc|cccc}
          \toprule
          \multirow{2}{*}{End-to-end Model} & \multirow{2}{*}{VLM} & \multicolumn{4}{c|}{\textbf{Detection}} & \multicolumn{4}{c|}{\textbf{Tracking}} & \multicolumn{4}{c|}{\textbf{Mapping}} & \multicolumn{4}{c}{\textbf{Motion Forecasting}}\\
          & & mAP $\uparrow$ & mATE $\downarrow$ & mASE $\downarrow$ & NDS $\uparrow$ & AMOTA $\uparrow$ & AMOTP $\downarrow$ & Recall $\uparrow$ & IDS $\downarrow$ & $AP_{ped}$ $\uparrow$ & $AP_{div}$ $\uparrow$ & $AP_{bou}$ $\uparrow$ & mAP $\uparrow$ & mADE $\downarrow$ & mFDE $\downarrow$ & MR $\downarrow$ & EPA $\uparrow$\\
         \hline
         \multirow{4}{*}{VAD} & None & 0.27 & 0.70 & 0.30 & 0.39 & - & - & - & - & 40.6 & 51.5 & 50.6 & 47.6 & 0.78 & 1.07 & 0.121 & 0.598\\
          & Llama-Adapter & 0.33 & 0.58 & 0.28 & 0.44 & - & - & - & - & 43.0 & 53.6 & 53.7 & 50.1 & 0.73 & 1.04 & 0.112 & 0.605\\
          & Llava-7B & 0.30 & 0.61 & 0.29 & 0.41 & - & - & - & - & 42.7 & 52.9 & 53.4 & 49.7 & 0.75 & 1.04 & 0.116 & 0.601\\
          & Qwen2.5-7B & 0.36 & 0.56 & 0.27 & 0.44 & - & - & - & - & 47.1 & 54.2 & 55.2 & 52.2 & 0.69 & 0.98 & 0.110 & 0.609\\
         \hline
         \multirow{4}{*}{SparseDrive} & None & 0.42 & 0.57 & 0.28 & 0.53 & 0.39 & 1.25  & 0.50 & 886 & 49.9 & 57.0 & 58.4 & 55.1 & 0.62 & 0.99 & 0.136 & 0.482\\
         & Llama-Adapter & 0.45 & 0.52 & 0.25 & 0.56 & 0.43 & 1.22 & 0.53 & 864 & 52.3 & 58.1 & 58.9 & 56.4 & 0.60 & 0.94 & 0.131 & 0.489\\
         & Llava-7B & 0.43 & 0.52 & 0.27 & 0.54 & 0.42 & 1.22 & 0.52 & 879 & 51.8 & 58.0 & 58.6 & 56.1 & 0.61 & 0.97 & 0.133 & 0.487\\
         & Qwen2.5-7B & 0.46 & 0.50 & 0.24 & 0.57 & 0.45 & 1.20 & 0.53 & 857 & 55.0 & 58.5 & 59.3 & 57.6 & 0.58 & 0.92 & 0.129 & 0.498\\    
         \bottomrule
    \end{tabular}%
    }
\label{tab:nuscenes-per}
\end{table*}


In Tab. \ref{tab:planning}, we compare various end-to-end driving configurations incorporating different VLAs against the original end-to-end models for both open-loop and closed-loop planning. For open-loop evaluation, following \cite{li2024ego,song2025momad}, we deactivate the ego status and adopt the evaluation protocols from UniAD \cite{hu2023uniad}. Our results demonstrate that integrating end-to-end models with VLMs and NavigScene substantially enhances performance, with Qwen2.5-7B yielding particularly significant improvements in both L2 and collision rate metrics. For closed-loop evaluation, we assess planning performance using multiple metrics: no at-fault collision (NC), drivable area compliance (DAC), time-to-collision (TTC), comfort (Comf.), ego progress subscores (EP), and the predictive driver model score (PDMS), all reported as percentages. Integrating VLMs with NavigScene significantly improves system performance, particularly in DAC, EP, and PDMS metrics, which correlate strongly with human-like driving capability and accurate navigation interpretation. These results highlight the importance of incorporating BVR knowledge in closed-loop planning. These results are based on DriveLM, with additional results using NuInstruct available in the \textcolor{blue}{SM}.

In Tab. \ref{tab:nuscenes-per}, we compare various end-to-end driving configurations across tasks including detection, tracking, mapping, and motion forecasting. We observe that NavigScene enhances the driving system's performance even in non-planning tasks such as detection. Notably, in detection mAP with Qwen2.5-7B, we achieve improvements of 0.09 over VAD and 0.04 over SparseDrive. Additional results using NuInstruct are available in the \textcolor{blue}{SM}.

\subsection{Qualitative Results on End-to-end Driving}

\begin{figure*}[!ht]
    \centering
    \includegraphics[width=1.0\linewidth]{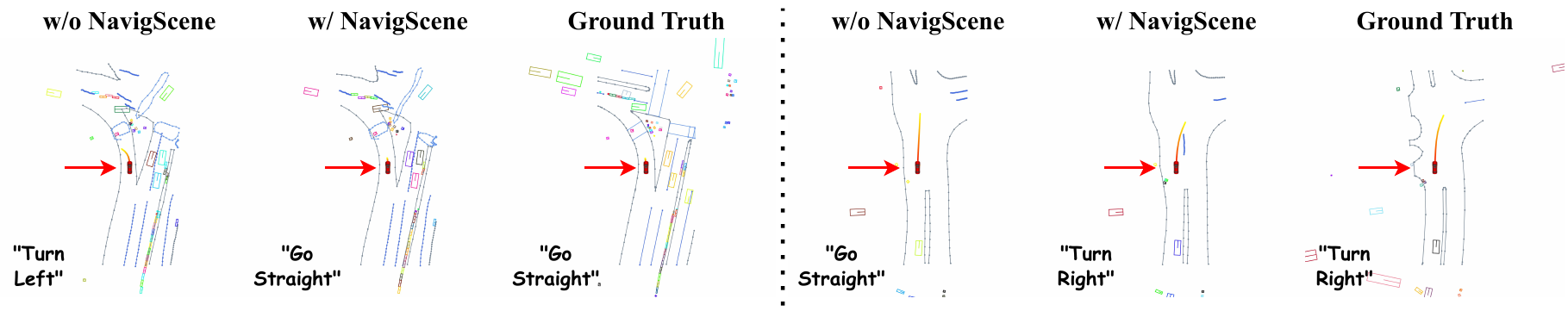}
    \caption{BEV visualization of open-loop planning on nuScenes dataset. ``w/o NavigScene'' indicates the original SparseDrive, and ``w/ NavigScene'' is the VLA model integrates Qwen2.5-7B (post-trained on \textcolor{blue}{DriveLM} and NavigScene via NSFT and NPO) with SparseDrive. Arrows point at ego vehicles, and text in bottom-left displays predicted driving commands, and orange curves represent predicted driving routes. Left: Vehicle proceeds straight and reduces speed to stop. Right: Vehicle anticipates a right turn by initiating an early lane change.}
    \label{fig:open-vis}
\end{figure*}

In Fig. \ref{fig:open-vis}, we present two open-loop planning examples comparing performance with and without NavigScene integration. Leveraging beyond-view-range knowledge from NavigScene enables the autonomous driving system to generate more accurate driving commands and route planning. This is particularly evident in the right case, where the vehicle correctly anticipates a right turn by initiating an early lane change with the assistance of global navigation guidance. Additional examples are exhibited in the \textcolor{blue}{SM}.

\begin{figure*}[!ht]
    \centering
    \includegraphics[width=1.0\linewidth]{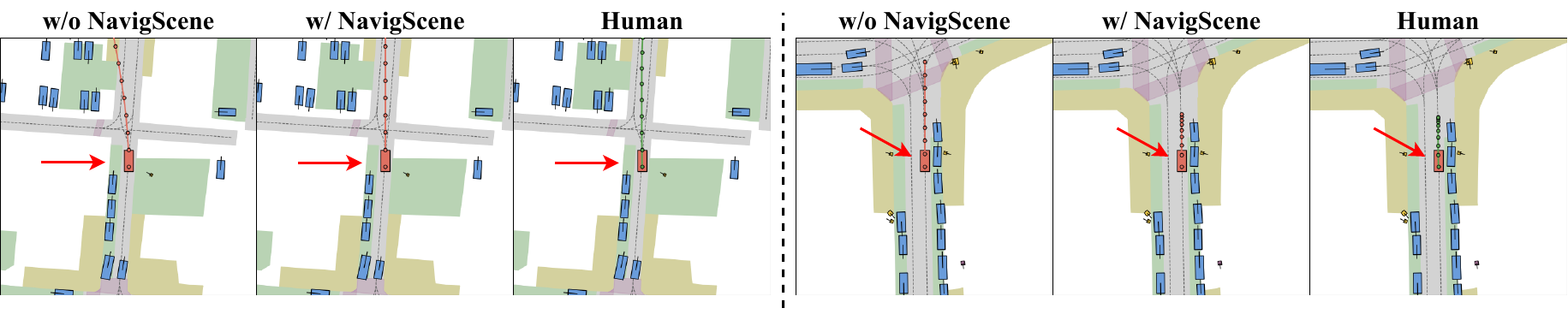}
    \caption{BEV Visualization of closed-loop planning on NAVSIM dataset. w/o NavigScene indicates the original SparseDrive, and w/ NavigScene is the VLA model integrates Qwen2.5-7B (post-trained on \textcolor{blue}{DriveLM} and NavigScene via NSFT and NPO) with SparseDrive. Arrows point at ego vehicles, with orange routes generated by models and green routes showing human operations. Left: Vehicle anticipates a left turn by initiating an early lane change. Right: Vehicle anticipates a right turn by slowing down and waiting.}
    \label{fig:close-vis}
\end{figure*}

In Fig. \ref{fig:close-vis}, we present two closed-loop planning examples demonstrating the impact of NavigScene integration. In the left example, the vehicle intends to switch to the fast track but should continue straight. Without navigation guidance, the driving model incorrectly turns left. Similarly, in the right example, the vehicle needs to turn right at the upcoming intersection. Without NavigScene, it continues straight for an extended distance, whereas with NavigScene, it appropriately slows down and waits for an opportunity to turn right. Extra examples are shown in the \textcolor{blue}{SM}.

\subsection{Ablation Study on Q\&A Tasks}

\begin{table}[!ht]
    \scriptsize
    \centering
    \caption{Ablation study on DriveLM-nuScenes}
    \resizebox{1.0\linewidth}{!}{%
    \begin{tabular}{c|c c|ccccccc}
          \toprule
          VLM & NSFT & NPO &BLEU-4 $\uparrow$ & METEOR $\uparrow$ & CIDEr $\uparrow$ & ROUGE\_L $\uparrow$ & SPICE $\uparrow$ & GPT $\uparrow$ & Comp. $\uparrow$ \\
         \hline
         \multirow{4}{*}{Llama-Adapter} & $\times$ & $\times$ & 50.68 & 33.75  & 2.37 & 64.59 & 44.20 & 70.83 & 29.98\\
          & $\checkmark$ & $\times$ & 52.20 & 35.46 & 2.45 & 66.53 & 45.79 & 72.18 & 31.74\\
          & $\times$ & $\checkmark$ & 51.73 & 34.89 & 2.53 & 66.20 & 46.38 & 71.46 & 31.44\\
          & $\checkmark$ & $\checkmark$ & 54.25 & 37.62 & 2.81 & 67.66 & 48.35 & 74.08 & 33.07 \\
         \hline
         \multirow{4}{*}{Llava-7B} & $\times$ & $\times$ & 49.75 & 33.21  & 2.19 & 63.84 & 42.56 & 70.24 & 29.37 \\
         & $\checkmark$ & $\times$ & 51.66 & 34.54 & 2.41 & 65.28 & 44.15 & 71.59 & 30.62 \\
         & $\times$ & $\checkmark$ & 50.84 & 34.07 & 2.35 & 64.86 & 44.01 & 71.19 & 30.46\\
         & $\checkmark$ & $\checkmark$ & 53.93 & 36.86 & 2.75 & 66.97 & 46.83 & 73.77 & 32.79\\
         \hline
         \multirow{4}{*}{Qwen2.5-7B} & $\times$ & $\times$ & 51.65 & 34.12  & 2.46 & 64.97 & 46.45 & 71.29 & 30.31\\
         & $\checkmark$ & $\times$ & 53.47 & 35.95 & 2.77 & 66.23 & 47.84 & 73.15 & 32.08\\
         & $\times$ & $\checkmark$ & 52.97 & 36.04 & 2.82 & 65.55 & 47.36 & 73.08 & 31.99\\
         & $\checkmark$ & $\checkmark$ &  55.13 & 38.20 & 3.14 & 67.88 & 49.89 & 74.87 & 34.26\\    
         \bottomrule
    \end{tabular}%
    }
    \vspace{-5pt}
\label{tab:drivelm-ab}
\end{table}

\begin{table*}[!ht]
    \scriptsize
    \centering
    \caption{Ablation study on NuInstruct}
    \resizebox{1.0\linewidth}{!}{%
    \begin{tabular}{c|c c|cccccc|cc|cccccc|c}
          \toprule
          \multirow{2}{*}{VLM} & \multirow{2}{*}{NSFT} & \multirow{2}{*}{NPO} & \multicolumn{6}{c|}{\textbf{Perception}} & \multicolumn{2}{c|}{\textbf{Prediction}} & \multicolumn{6}{c|}{\textbf{Risk}} & \multirow{2}{*}{\textbf{Planning} $\uparrow$} \\
          & & & Dis $\downarrow$ & Spe $\downarrow$ & Ins $\downarrow$ & Clo $\uparrow$ & Sta $\uparrow$ & SaR $\uparrow$ & Mot $\downarrow$ & Sta $\uparrow$ & App $\uparrow$ & Lan $\uparrow$ & Onc $\uparrow$ & Cro $\uparrow$ & Ove $\uparrow$ & Bra $\uparrow$ & \\
         \hline
         \multirow{4}{*}{Llama-Adapter} & $\times$ & $\times$ & 27.9 & 6.6 & 6.5 & 20.4 & 16.2 & 24.0 & 8.7 & 40.2 & 13.5 & 18.7 & 16.2 & 18.4 & 6.5 & 21.5 & 25.7 \\
          & $\checkmark$ & $\times$ & 25.8 & 5.3 & 5.5 & 24.9 & 17.4 & 26.1 & 7.2 & 41.8 & 14.4 & 20.1 & 17.9 & 20.7 & 7.6 & 23.9 & 27.7\\
          & $\times$ & $\checkmark$ & 26.0 & 5.5 & 5.7 & 24.1 & 17.0 & 25.8 & 7.5 & 42.2 & 14.6 & 19.2 & 17.3 & 19.9 & 7.5 & 22.8 & 27.4 \\
          & $\checkmark$ & $\checkmark$ & 24.3 & 3.9 & 4.0 & 32.2 & 19.8 & 28.7 & 4.3 & 44.1 & 15.8 & 21.0 & 19.9 & 22.8 & 9.0 & 26.4 & 31.2 \\
         \hline
         \multirow{4}{*}{Llava-7B}  & $\times$ & $\times$ & 28.4 & 6.9 & 6.6 & 22.1 & 16.2 & 23.5 & 9.4 & 38.9 & 12.3 & 19.6 & 16.9 & 18.7 & 6.5 & 21.3 & 25.3 \\
         & $\checkmark$ & $\times$ & 26.3 & 5.8 & 5.4 & 24.7 & 17.5 & 25.1 & 8.0 & 40.4 & 13.6 & 20.5 & 18.0 & 20.4 & 7.7 & 23.0 & 27.2\\
         & $\times$ & $\checkmark$ & 26.9 & 5.5 & 5.1 & 24.4 & 17.1 & 24.7 & 8.6 & 39.9 & 13.8 & 20.7 & 18.5 & 20.0 & 7.3 & 23.2 & 27.0\\
          & $\checkmark$ & $\checkmark$ & 24.5 & 4.3 & 4.1 & 27.6 & 19.5 & 27.8 & 6.3 & 43.6 & 15.0 & 22.7 & 20.2 & 23.0 & 9.2 & 26.8 & 32.6 \\
         \hline
         \multirow{4}{*}{Qwen2.5-7B} & $\times$ & $\times$ & 26.9 & 5.6 & 5.4 & 26.5 & 17.3 & 26.7 & 7.8 & 41.8 & 15.2 & 20.3 & 18.4 & 20.5 & 7.7 & 22.1 & 26.6 \\
         & $\checkmark$ & $\times$ & 25.2 & 4.4 & 4.1 & 29.8 & 18.9 & 28.0 & 6.5 & 43.7 & 17.1 & 21.9 & 20.3 & 23.2 & 8.4 & 24.5 & 29.9 \\
         & $\times$ & $\checkmark$ & 25.5 & 4.8 & 4.5 & 29.3 & 18.4 & 28.3 & 6.9 & 43.2 & 16.8 & 21.5 & 20.4 & 23.4 & 8.1 & 24.2 & 29.7\\
          & $\checkmark$ & $\checkmark$ & 23.6 & 3.1 & 3.2 & 33.7 & 20.2 & 31.9 & 3.8 & 45.3 & 18.6 & 23.7 & 21.9 & 26.2 & 10.4 & 27.9 & 36.4 \\    
         \bottomrule
    \end{tabular}%
    }
\label{tab:NuInstruct-ab}
\end{table*}

In Tables \ref{tab:drivelm-ab} and \ref{tab:NuInstruct-ab}, we present ablation studies conducted on DriveLM-nuScenes and NuInstruct datasets, respectively. We examine four experimental settings: (1) without NSFT and NPO, where the VLM is finetuned without NavigScene; (2) with NSFT only, where the VLM undergoes fine-tuning with NavigScene; (3) with NPO only, where the VLM is first finetuned without NavigScene and then trained using NPO and NavigScene; and (4) with both NSFT and NPO, where the VLM is first finetuned with NavigScene and subsequently trained with NPO and NavigScene.

Based on the tables provided, both the DriveLM-nuScenes and NuInstruct datasets demonstrate consistent performance improvements when applying NSFT and NPO across all three VLMs. The most significant performance gains occur when both NSFT and NPO are applied together, with Qwen2.5-7B showing the highest overall scores across metrics in both datasets. For DriveLM-nuScenes, the combination of NSFT and NPO improved BLEU-4, METEOR, and CIDEr substantially, with Qwen2.5-7B achieving the best CIDEr score of 3.14. Similarly, in the NuInstruct dataset, this combined approach led to strong reductions in Dis and Spe. These results suggest that the collaboration of NSFT and NPO techniques enhances VLMs' ability to understand complex driving scenarios and generate appropriate responses.

\subsection{Ablation Study on Open-loop and Closed-loop Planning}

\begin{table*}[!ht]
    \scriptsize
    \centering
    \caption{Ablation study on both open-loop and closed-loop planning settings. Left: Open-loop planning performance on nuScenes. Right: Closed-loop planning performance on NAVSIM. All VLMs' NSFT and NPO are based on \textcolor{blue}{DriveLM-nuScenes} and NavigScene-nuScenes.}
    \resizebox{1.0\linewidth}{!}{%
    \begin{tabular}{c|c|cc|cccc|cccc|ccccc|c}
          \toprule
          \multirow{3}{*}{\begin{tabular}[c]{@{}c@{}}End-to-end\\Model\end{tabular}} & \multirow{3}{*}{VLM} & \multirow{3}{*}{NSFT} & \multirow{3}{*}{NPO} & \multicolumn{8}{c|}{\textbf{Open-loop Planning on nuScenes}} & \multicolumn{6}{c}{\textbf{Closed-loop Planning on NAVSIM}} \\
          \cline{5-18}
          & & & & \multicolumn{4}{c|}{\textbf{L2 (m)$\downarrow$}} & \multicolumn{4}{c|}{\textbf{Collision Rate (\textpertenthousand) $\downarrow$}} & \multirow{2}{*}{NC $\uparrow$} & \multirow{2}{*}{DAC $\uparrow$} & \multirow{2}{*}{TTC $\uparrow$} & \multirow{2}{*}{Comf. $\uparrow$} & \multirow{2}{*}{EP $\uparrow$} & \multirow{2}{*}{PDMS $\uparrow$} \\
          \cline{5-12}
          & & & & 1s & 2s & 3s & Avg. & 1s & 2s & 3s & Avg. & & & & & & \\
         \hline
         \multirow{9}{*}{VAD} & None & N/A & N/A & 0.54 & 1.15 & 1.98 & 1.22 & 9.76 & 24.20 & 95.93 & 43.30 & 97.0 & 86.5 & 89.7 & 100 & 75.4 & 80.6 \\
         \cline{2-18}
         & \multirow{4}{*}{Llama-Adapter} & $\times$ & $\times$ & 0.69 & 1.48 & 2.64 & 1.60 & 11.35 & 28.64 & 104.57 & 48.19 & 95.6 & 84.2 & 87.1 & 99.8 & 72.5 & 78.2 \\
         & & $\checkmark$ & $\times$ & 0.50 & 1.07 & 1.88 & 1.15 & 8.53 & 22.84 & 89.77 & 40.38 & 97.3 & 87.7 & 90.2 & 100 & 76.6 & 81.3\\
         & & $\times$ & $\checkmark$ & 0.52 & 1.12 & 1.90 & 1.18 & 8.79 & 23.01 & 91.24 & 41.01 & 97.2 & 87.4 & 90.0 & 100 & 75.9 & 81.0\\
         & & $\checkmark$ & $\checkmark$ & 0.42 & 1.02 & 1.76 & 1.07 & 6.28 & 21.85 & 81.37 & 36.50 & 97.9 & 88.8 & 91.6 & 100 & 77.6 & 82.2 \\
         \cline{2-18}
         & \multirow{4}{*}{Qwen2.5-7B} & $\times$ & $\times$ & 0.66 & 1.50 & 2.58 & 1.58 & 12.48 & 29.96 & 107.25 & 49.90 & 95.3 & 84.4 & 86.6 & 99.5 & 72.8 & 77.8\\
         & & $\checkmark$ & $\times$ & 0.45 & 1.03 & 1.82 & 1.10 & 7.24 & 22.38 & 85.69 & 38.44 & 97.5 & 88.0 & 90.4 & 100 & 76.9 & 81.5\\
         & & $\times$ & $\checkmark$ & 0.49 & 1.06 & 1.85 & 1.12 & 7.85 & 22.46 & 87.33 & 39.21 & 97.6 & 88.4 & 91.1 & 100 & 77.2 & 81.9\\
         & & $\checkmark$ & $\checkmark$ & 0.37 & 0.96 & 1.65 & 0.99 & 4.52 & 18.10 & 74.62 & 32.41 & 98.3 & 90.6 & 92.4 & 100 & 79.0 & 84.0 \\
         \bottomrule
         \multirow{9}{*}{SparseDrive} & None & N/A & N/A & 0.44 & 0.92 & 1.69 & 1.01 & 7.38 & 19.46 & 70.60 & 32.48 & 97.2 & 91.7 & 91.4 & 100 & 77.9 & 82.4 \\
         \cline{2-18}
         & \multirow{4}{*}{Llama-Adapter} & $\times$ & $\times$ & 0.62 & 1.38 & 2.40 & 1.47 & 10.53 & 25.57 & 98.19 & 44.76 & 96.1 & 85.9 & 87.6 & 100 & 73.3 & 78.7 \\
         & & $\checkmark$ & $\times$ & 0.39 & 0.87 & 1.62 & 0.96 & 7.10 & 18.35 & 65.99 & 30.48 & 97.6 & 92.2 & 91.7 & 100 & 78.2 & 82.7\\
         & & $\times$ & $\checkmark$ & 0.42 & 0.89 & 1.66 & 0.99 & 7.25 & 18.67 & 68.23 & 31.38 & 97.6 & 92.6 & 92.3 & 100 & 78.6 & 82.9\\
         & & $\checkmark$ & $\checkmark$ & 0.33 & 0.78 & 1.47 & 0.86 & 5.97 & 15.65 & 58.41 & 26.68 & 98.0 & 93.5 & 93.0 & 100 & 80.4 & 83.9  \\
         \cline{2-18}
         & \multirow{4}{*}{Qwen2.5-7B} & $\times$ & $\times$ & 0.58 & 1.27 & 2.35 & 1.40 & 10.06 & 24.94 & 97.83 & 44.28 & 96.4 & 86.6 & 88.1 & 99.8 & 73.4 & 79.0\\
         & & $\checkmark$ & $\times$ & 0.38 & 0.87 & 1.60 & 0.95 & 6.97 & 17.84 & 64.41 & 29.74 & 97.8 & 92.3 & 92.0 & 100 & 79.0 & 83.6\\
         & & $\times$ & $\checkmark$  & 0.40 & 0.88 & 1.66 & 0.98 & 7.03 & 18.06 & 67.69 & 30.93 & 97.8 & 92.0 & 91.8 & 100 & 78.9 & 83.5 \\
         & & $\checkmark$ & $\checkmark$ & 0.29 & 0.64 & 1.35 & 0.76 & 4.38 & 12.99 & 44.75 & 20.71 & 98.3 & 96.0 & 94.1 & 100 & 81.7 & 86.5 \\
         \bottomrule
    \end{tabular}%
    }
\label{tab:planning-ab}
\end{table*}

In Tab. \ref{tab:planning-ab}, we present ablation studies conducted on nuScenes for open-loop planning and NAVSIM for closed-loop planning, respectively. We examine five experimental settings: (1) without VLM, where only the end-to-end model is applied; (2) without NSFT and NPO, where the VLM is finetuned without NavigScene; (3) with NSFT only, where the VLM undergoes fine-tuning with NavigScene; (4) with NPO only, where the VLM is first finetuned without NavigScene and then trained using NPO and NavigScene; and (5) with both NSFT and NPO, where the VLM is first finetuned with NavigScene and subsequently trained with NPO and NavigScene. All VLMs are frozen then connected with end-to-end models to construct a VLA model for autonomous driving.

Based on the ablation study in Table \ref{tab:planning-ab}, there are obvious performance improvements when incorporating NSFT and NPO techniques in both open-loop and closed-loop planning across both VAD and SparseDrive end-to-end models. We observe SparseDrive generally outperforms VAD, while Qwen2.5-7B consistently outperforms Llama-Adapter across all configurations. Notably, even the baseline systems without VLM integration ("None") perform reasonably well, but VLM integration with proper post-training techniques (NSFT+NPO) provides substantial gains, particularly in reducing collision rates and improving trajectory accuracy. The synergistic effect of combining NSFT and NPO demonstrates that navigation-specific fine-tuning followed by preference optimization creates the most effective autonomous driving systems.

\subsection{Cross-city Generalization on End-to-end Driving}

\begin{table}[!ht]
    \scriptsize
    \centering
    \caption{Cross-city generalization results on nuScenes. All VLMs were first conducted NSFT on \textcolor{blue}{DriveLM-nuScenes} and NavigScene-nuScenes.}
    \resizebox{1.0\linewidth}{!}{%
    \begin{tabular}{c|c|c|cc|cc}
          \toprule
          \multirow{2}{*}{End-to-end Model} & \multirow{2}{*}{VLM} & \multirow{2}{*}{NPO} & \multicolumn{2}{c|}{\textbf{Boston} $\rightarrow$ \textbf{Singapore} } & \multicolumn{2}{c}{\textbf{Singapore} $\rightarrow$ \textbf{Boston}} \\
          & & & Avg. L2 (m) $\downarrow$ & Avg. Col (\textpertenthousand) $\downarrow$ & Avg. L2 (m) $\downarrow$ & Avg. Col (\textpertenthousand) $\downarrow$ \\
         \hline
         \multirow{5}{*}{VAD} & None & N/A & 0.86 & 26.83 & 0.63 & 20.44 \\
          \cline{2-7}
          & \multirow{2}{*}{Llama-Adapter} & $\times$ & 0.94 & 27.08 & 0.75 & 21.19 \\
          & & $\checkmark$ & 0.75 & 23.29 & 0.64 & 19.30\\
          \cline{2-7}
          & \multirow{2}{*}{Qwen2.5-7B} & $\times$ & 0.97 & 27.51 & 0.81 & 21.85\\
          & & $\checkmark$ & 0.70 & 22.55 & 0.61 & 18.46 \\ 
         \bottomrule
         \multirow{5}{*}{SparseDrive} & None & N/A & 0.97 & 30.17   & 0.84 & 33.62\\
          \cline{2-7}
          & \multirow{2}{*}{Llama-Adapter} & $\times$ & 1.04 & 27.84 & 0.87 & 34.48 \\
          & & $\checkmark$ & 0.88 & 27.32 & 0.72 & 20.99 \\
          \cline{2-7}
          & \multirow{2}{*}{Qwen2.5-7B} & $\times$ & 1.11 & 28.06 & 0.93 & 35.64\\
          & & $\checkmark$ & 0.82 & 24.70 & 0.69 & 19.66\\
         \bottomrule
    \end{tabular}%
    }
\label{tab:gen}
\end{table}
\vspace{-3pt}

In Table \ref{tab:gen}, we present cross-city generalization results on the two cities from nuScenes dataset. Following \cite{pan2024vlp}, we examine two transfer tasks: Boston $\rightarrow$ Singapore and Singapore $\rightarrow$ Boston. In the first task, models are trained on Boston data and evaluated in a zero-shot manner on Singapore data, while in the second task, this process is reversed. Additional results are shown in the \textcolor{blue}{SM}.

The cross-city generalization results demonstrate the significant impact of NPO on autonomous driving systems' generalization ability. When testing models trained in Boston on Singapore data and vice versa, VLA models with NPO consistently outperform both original end-to-end architectures (VAD and SparseDrive) and VLA models without NPO. 
These results highlight NPO's effectiveness in enhancing autonomous driving systems' robustness when navigating unfamiliar urban environments with different traffic patterns and infrastructure designs.

\section{Conclusion}

In this paper, we address a critical limitation in current autonomous driving systems: the disconnection between local sensor data and global navigation context. First we introduced NavigScene, an auxiliary navigation-guided natural language dataset that bridges this gap by simulating \emph{human-like} driving environments. Besides, through three complementary paradigms based on NavigScene: Navigation-guided Reasoning, Navigation-guided Preference Optimization, and Navigation-guided Vision-Language-Action model, we achieve significant improvements in driving-related tasks across Q\&A, perception, prediction, and planning. We enable reasoning capability beyond visual range and enhance generalization ability to diverse driving scenarios. While our work represents a significant step toward more comprehensive autonomous driving systems, future research should focus on integrating dynamic and multi-modal navigation information. In a word, this work brings autonomous driving systems closer to \emph{human-like} ability to navigate complex, unfamiliar environments with improved reliability and safety.

\newpage

\bibliographystyle{ACM-Reference-Format}
\bibliography{sample-base}


\end{document}